\documentclass[10pt,twocolumn,letterpaper]{article}

\usepackage[pagenumbers]{cvpr} 

\usepackage{amsthm}
\usepackage{multirow}
\usepackage{array}
\usepackage{tabularx}
\usepackage{bm}
\usepackage{makecell}

\theoremstyle{plain}

\theoremstyle{remark}

\newcommand{\ours}{RS-Avatar\xspace}
\newcommand{\dsyn}{RS-ZJU\xspace}

\newcommand{\R}{\mathbb{R}}
\newcommand{\SEthree}{\mathrm{SE}(3)}
\newcommand{\SOthree}{\mathrm{SO}(3)}
\newcommand{\bx}{\mathbf{x}}

\newcommand{\bT}{\mathbf{T}}
\newcommand{\bR}{\mathbf{R}}
\newcommand{\bt}{\mathbf{t}}
\newcommand{\bG}{\mathcal{G}}
\newcommand{\bS}{\mathcal{S}}
\newcommand{\bW}{\mathcal{W}}
\newcommand{\bL}{\mathcal{L}}
\newcommand{\bP}{\Pi}
\newcommand{\rnd}{\mathcal{R}}

\renewcommand{\paragraph}[1]{\vspace{.4em}\noindent\textbf{#1.}}

\definecolor{cvprblue}{rgb}{0.21,0.49,0.74}
\usepackage[pagebackref,breaklinks,colorlinks,allcolors=cvprblue]{hyperref}

\def\paperID{*****} 
\def\confName{CVPR}
\def\confYear{2026}

\title{Scanline-Aware Animatable Gaussian Avatars from Rolling-Shutter Videos}

\author{
Youxiang Wang \\
University of Macau \\
}

\begin{document}
\maketitle
\begin{abstract}
Animatable human avatars are routinely reconstructed from multi-view video under a silent assumption: that every pixel of a frame observes the same instant of the body's motion. Rolling-shutter (RS) sensors expose image rows sequentially, so within one frame the head and the feet of a moving person are separated by tens of milliseconds of articulated motion, and \emph{every scanline sees a different pose}. Feeding such video to a state-of-the-art avatar bakes the distortion into the canonical representation, where it survives as shear and wobble under novel views and novel poses. Worse, every camera in a rig follows its own readout schedule, so the multi-view consistency that drives the reconstruction is violated even when the geometry is correct. We present \ours, which reconstructs a sharp, undistorted, animatable 3D Gaussian avatar directly from RS video. The formulation is minimal: a motion-aware avatar already renders the body at several sub-frame instants, and where a blur model \emph{averages} those renderings, a rolling-shutter model \emph{composites} them scanline by scanline. Changing that operator is the only modification required. On \dsyn, a benchmark we build from ZJU-MoCap, this improves novel-view synthesis over training as if the frames were instantaneous, on every subject. A motion-aware \emph{blur} model built on the same sub-frame machinery does not transfer, and in fact falls below the shutter-oblivious baseline: the machinery is reusable, the operator is not.
\end{abstract}

\section{Introduction}
\label{sec:intro}

Reconstructing an animatable 3D human avatar from multi-view video is a cornerstone problem in vision and graphics, with applications spanning telepresence, film production, sports analysis and embodied AI.
Driven by neural radiance fields~\cite{mildenhall2020nerf} and, more recently, by 3D Gaussian splatting~\cite{kerbl20233dgs}, a mature pipeline has emerged: a canonical set of primitives is bound to a parametric body model such as SMPL~\cite{loper2015smpl} and warped into each observed frame by linear blend skinning, so that a handful of minutes of video yields a drivable, photorealistic avatar~\cite{peng2021neuralbody,peng2021animatable,weng2022humannerf,jiang2023instantavatar,hu2024gauhuman,qian20243dgsavatar,li2024animatablegaussians,hu2024gaussianavatar}.

Every method in this lineage rests on an assumption that is rarely stated and almost never satisfied: that an image is an instantaneous snapshot, so all of its pixels can be explained by a \emph{single} body pose.
Real sensors disagree.
The overwhelming majority of CMOS cameras---from smartphones and action cameras to the machine-vision modules used in capture studios---employ a \emph{rolling shutter} (RS), reading out image rows sequentially rather than simultaneously.
The readout of a full frame typically occupies a large fraction of the frame period; for a $30$\,fps sensor with a $30$\,ms readout, the last row of a frame is captured almost $30$\,ms after the first.
A wrist swinging at $5$\,m/s therefore travels $15$\,cm between the top and the bottom of a single image.
For a static scene this manifests as the familiar skew of a passing car; for an articulated human body, it is far worse, because different body parts occupy different rows and are thus sampled at different times, and because the distortion is \emph{non-rigid} and changes with the pose itself.

Confronted with RS input, one can ignore the problem or reach for one of three families of existing tools. None of the four works, and the reasons differ.

\paragraph{Ignoring the shutter}
Training a state-of-the-art avatar directly on RS frames does not merely blur the result---it corrupts the canonical model. The optimizer must explain a scanline-dependent geometry with a single pose per frame, and the only degrees of freedom it has are the canonical primitives themselves, so the shear is absorbed into canonical space. Once there, it reappears whenever the avatar is rendered from a novel view or driven by a novel pose.
The corruption is also \emph{view-dependent}: two cameras looking at the same body at the same nominal timestamp read out different rows at different instants, so their observations are mutually inconsistent even when the geometry is perfect.
This violates the multi-view photometric consistency that drives the reconstruction, and the optimizer resolves the conflict by over-smoothing and thickening limbs.

\paragraph{Correcting the images first}
A two-stage pipeline that runs an RS correction network~\cite{liu2020dsun,fan2021sunet,fan2021rssr,fan2022cvr,zhong2022dualrs,shang2023selfdrsc,cao2024intermediate} before avatar training inherits the weaknesses of monocular 2D correction.
These methods estimate a per-pixel distortion flow from a single view (or a dual-reversed pair), which is severely ill-posed for a non-rigidly articulating body; they hallucinate content in disoccluded regions; and, critically, they correct each camera independently, so the corrected views are not guaranteed to agree in 3D.
Errors introduced in stage one are indistinguishable from appearance in stage two.

\paragraph{Borrowing RS-aware neural rendering}
A recent line of work makes radiance fields and Gaussian splatting robust to rolling shutter~\cite{niu2024rsnerf,li2024usbnerf,xu2024ursnerf,seiskari2024gsmove}.
All of it addresses the \emph{dual} of our problem: a \emph{static} scene observed by a \emph{moving} camera, where the RS effect is absorbed into a six-degree-of-freedom camera trajectory that is optimized alongside the scene.
Our setting inverts this.
The cameras are rigidly mounted and calibrated; it is the \emph{subject} that moves, deforms non-rigidly and self-occludes.
No camera trajectory, however finely parameterized, can express a person waving one arm while standing still---the required deformation lives in the articulated pose space of the body, not in $\SEthree$.

\paragraph{Reusing the blur formulation}
Closest in spirit is recent work on reconstructing avatars from \emph{blurry} video~\cite{niu2026madavatar}, which also couples a body-motion model to the image formation process.
But blur and rolling shutter are structurally different corruptions, and the difference lands precisely where the image formation model is written down.
Blur is a temporal \emph{integration}: it averages the appearance over an exposure, discarding high-frequency radiometric content and leaving a large null space.
Rolling shutter, in the short-exposure regime, discards nothing---it is a measure-preserving \emph{re-indexing} that assigns each row a different point on the motion trajectory.
Consequently their ambiguities are opposites.
The blur operator is invariant to time reversal within an exposure, so the \emph{direction} of motion is unidentifiable, which is exactly what the blur-avatar formulation must regularize away.
The RS operator fixes the arrow of time by construction and has no such ambiguity.
So the blur formulation does not transfer---but, encouragingly, its \emph{scaffolding} does.

\medskip
\ours{} is built on this observation.
A motion-aware avatar already does the expensive part of the job: it recovers a sub-frame body-motion model and renders the avatar at several instants inside a single frame.
What distinguishes the two corruptions is only how those renderings are combined into the synthesized observation.
Blur averages them uniformly; a rolling shutter selects among them per scanline, taking row band $i$ of the synthesized frame from the rendering at the time that band was read out.
Replacing the uniform average $\frac{1}{T}\sum_i$ with the masked sum $\sum_i \mathbf{M}_i\odot$ is, to a first approximation, the entire algorithmic change.

The two settings are not equivalent, though.
The inter-frame motion prior is left with a narrower role: in the blur setting it must also fix the direction of sub-frame motion, which the scan order here supplies for free, so what remains for it is continuity alone.
The temporal discretization also changes character, since band composition leaves structured seams where averaging leaves unstructured softening.
That turns out not to require more sub-frame renderings: sweeping the number of row bands over a factor of four does not improve the reconstruction, because the intra-frame trajectory is a low-order polynomial that a coarse discretization already oversamples.

We validate \ours{} on \dsyn, a new benchmark built by scanline-compositing temporally interpolated ZJU-MoCap~\cite{peng2021neuralbody} sequences, with the original sharp frames as ground truth.
Modeling the shutter improves the reconstruction over ignoring it, on every subject.
The more informative outcome concerns the blur model: with the same spline, the same $T$ renderings and the same optimizer, changing only the operator drops it \emph{below} the baseline that models no sub-frame structure at all.

Our contributions are:
\begin{enumerate}[leftmargin=*,itemsep=1pt,topsep=2pt]
  \item We introduce the problem of animatable avatar reconstruction from rolling-shutter video, and give a 3D-aware RS image formation model for animatable Gaussian avatars that differs from the blur model only in its combination operator.
  \item We identify what does \emph{not} carry over: the inter-frame motion prior loses one of its two roles, because the direction of sub-frame motion, which it must disambiguate under blur, is fixed for free by the scan order. We also show what one might expect to carry over but does not---the structured band-composition residual does \emph{not} demand a finer temporal discretization, because the sub-frame motion model, not the band count, is what bounds fidelity.
  \item We contribute \dsyn, a benchmark for the task, and evaluate against shutter-oblivious, two-stage-correction and blur-model baselines.
\end{enumerate}

\section{Related Work}
\label{sec:related}

\subsection{Animatable Human Avatars}
Modern avatar reconstruction couples a photorealistic appearance representation to a parametric body model.
Following NeRF~\cite{mildenhall2020nerf}, Neural Body~\cite{peng2021neuralbody} anchored latent codes on SMPL~\cite{loper2015smpl} vertices, Animatable NeRF~\cite{peng2021animatable} learned neural blend-weight fields, and HumanNeRF~\cite{weng2022humannerf} demonstrated free-viewpoint rendering from monocular video; InstantAvatar~\cite{jiang2023instantavatar} brought training time to minutes.
The advent of 3D Gaussian splatting~\cite{kerbl20233dgs} shifted the field to explicit primitives: GauHuman~\cite{hu2024gauhuman} binds Gaussians to SMPL and refines skinning weights, 3DGS-Avatar~\cite{qian20243dgsavatar} adds a non-rigid deformation field, and Animatable Gaussians~\cite{li2024animatablegaussians} and GaussianAvatar~\cite{hu2024gaussianavatar} learn pose-dependent Gaussian maps for high-fidelity clothing dynamics.
We adopt this canonical-plus-skinning design, but replace its instantaneous-capture assumption with an explicit shutter model.
Crucially, none of these methods can be repaired by simply supplying better per-frame poses: with a rolling shutter, \emph{no} single pose per frame is correct.

\subsection{Reconstruction from Degraded Observations}
In practice, many degradations exists when we want to perform 3D reconstruction, such as low-light~\cite{niu2023visibility,niu2023physics,niu2023nir,niu2026toward,guo2020zero,jiang2021enlightengan}, blurriness~\cite{ma2022deblurnerf,wang2023badnerf,zhao2024badgaussians,niu2026madavatar}, and rolling shutter~\cite{niu2024rsnerf,li2024usbnerf,xu2024ursnerf}.
A growing body of work relaxes the assumption that inputs are clean.
Deblur-NeRF~\cite{ma2022deblurnerf} learns per-view blur kernels; BAD-NeRF~\cite{wang2023badnerf} and BAD-Gaussians~\cite{zhao2024badgaussians} model the camera trajectory during exposure and bundle-adjust it jointly with the radiance field.
For humans, MAD-Avatar~\cite{niu2026madavatar} reconstructs sharp animatable Gaussian avatars from blurry multi-view video by modeling motion-induced blur as an average over sub-frame renderings driven by a per-frame pose spline, and resolves the resulting direction-of-motion ambiguity with an inter-frame continuity prior.
We deliberately inherit that scaffolding.
What we change is the operator that combines the sub-frame renderings (\cref{sec:formation}), and what we must then re-derive is everything that operator touches: the accuracy of the temporal discretization and the purpose of the continuity prior.

\subsection{Rolling Shutter Geometry and Correction}
Rolling shutter has a long history in geometric vision.
Ait-Aider~\etal~\cite{aitaider2006simultaneous} recovered object pose \emph{and} velocity from a single RS view, an early statement of how tightly the shutter couples motion to geometry.
RS-aware bundle adjustment~\cite{hedborg2012rolling}, stereo~\cite{saurer2013rolling}, differential SfM~\cite{zhuang2017rolling} and calibration-free video stabilization~\cite{grundmann2012calibration} established that the shutter must enter the geometric model, while continuous-time trajectory representations~\cite{lovegrove2013spline,sommer2020efficient} became the standard vehicle for doing so.
We use the same continuous-time idea, but place the spline in articulated body-pose space rather than in $\SEthree$ camera space.

A parallel line learns to undistort RS images directly: DSUN~\cite{liu2020dsun}, SUNet~\cite{fan2021sunet}, RSSR~\cite{fan2021rssr} and CVR~\cite{fan2022cvr} predict pixel-wise displacement or intermediate flow from consecutive RS frames; JCD~\cite{zhong2021jcd} jointly handles RS and blur; dual-reversed-distortion methods~\cite{zhong2022dualrs,shang2023selfdrsc} exploit two sensors scanning in opposite directions; and recent work improves robustness to nonlinear motion and occlusion~\cite{qu2023nonlinear,cao2024intermediate}.
These operate in 2D and per view, and therefore cannot enforce---or even represent---the 3D consistency that avatar reconstruction requires.

\subsection{Rolling-Shutter-Aware Neural Rendering}
Most closely related in machinery, RS-NeRF~\cite{niu2024rsnerf}, USB-NeRF~\cite{li2024usbnerf} and URS-NeRF~\cite{xu2024ursnerf} embed a row-dependent camera pose into volumetric rendering and bundle-adjust it with the radiance field, and Gaussian Splatting on the Move~\cite{seiskari2024gsmove} does the same for 3DGS using velocities from visual-inertial odometry.
Every one of these methods assumes a rigid, static scene: the entire time dependence is carried by a camera trajectory, and the scene representation itself is time-invariant.
Our subject is the opposite---a static, calibrated rig observing a non-rigidly deforming articulated body---so the time dependence must live in the \emph{deformation}, which changes the rendering problem (each primitive, not each ray, acquires its own timestamp), the parameterization (pose space, not $\SEthree$), and the degeneracy analysis.
To the best of our knowledge, \ours{} is the first method to reconstruct animatable human avatars from rolling-shutter video.

\section{Method}
\label{sec:method}

\begin{figure*}[t]
  \centering
  \includegraphics[width=\textwidth]{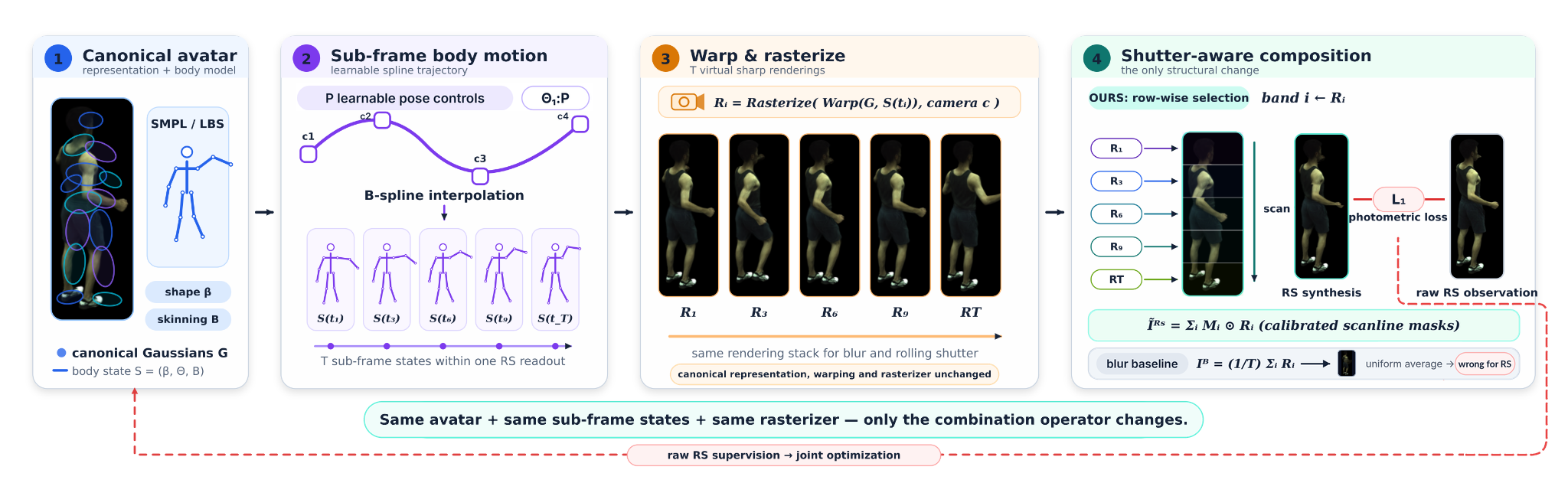}
  \caption{\textbf{Overview of \ours.}
  Per-frame spline control points give $T$ sub-frame body states; the canonical Gaussians are warped to each and rasterized into the stack $R_1\dots R_T$ (stages 1--3).
  Stage 4 is the method: band $i$ of the synthesized frame is taken from the rendering at the instant that band was read out, so the observation \emph{selects} among the renderings where the blur model beneath it averages them.
  The raw RS frames are the only supervision.}
  \label{fig:overview}
\end{figure*}

\subsection{Preliminaries}
\label{sec:prelim}

\paragraph{Animatable Gaussian avatar}
We represent the subject by $M$ Gaussian primitives in a canonical (T-pose) space,
$\bG=\{\bG_k\}_{k=1}^{M}$ with $\bG_k=(\bx_k,\mathbf{q}_k,\mathbf{s}_k,\alpha_k,\mathbf{c}_k)$ denoting position, rotation, scale, opacity and view-dependent color~\cite{kerbl20233dgs}.
A body state $\bS=(\beta,\Theta,\mathcal{B})$ collects the SMPL~\cite{loper2015smpl} shape coefficients $\beta\in\R^{10}$, the pose $\Theta$ (local rotations of $J\!=\!24$ joints plus global orientation and translation), and the skinning weights $\mathcal{B}$.
A canonical primitive is warped to observation space by linear blend skinning,
\begin{equation}
  \bW(\bx_k;\bS) \;=\; \bR\Big(\textstyle\sum_{j=1}^{J} w_{kj}\, \bT_j(\Theta,\beta)\,\bx_k\Big) + \bt,
  \label{eq:lbs}
\end{equation}
with the covariance transported by the rotational part of the same transform.
We write $\rnd\big(\bW(\bG,\bS);\bP_c\big)$ for the differentiable rasterization of the deformed primitives into camera $c$.

\paragraph{Capture setup}
$C$ calibrated cameras of height $H$ pixels observe the subject for $N$ frames at period $\Delta$.
Camera $c$ has readout time $\rho_c$ (the interval between the start of the first and of the last row) and trigger offset $\delta_c$, both obtained by standard sensor calibration.
We write $\gamma=\rho/\Delta$ for the \emph{readout ratio}, the fraction of the frame period spent scanning.

\subsection{Rolling-Shutter Image Formation}
\label{sec:formation}

\paragraph{From blur to rolling shutter}
Under a global shutter with a finite exposure, a moving subject produces a blurry image, classically modeled as the temporal integral of the latent sharp images over the exposure window~\cite{niu2026madavatar},
\begin{equation}
  I^{B}(u,v) \;=\; \frac{1}{\tau}\int_{0}^{\tau} I^{S}_{t}(u,v)\,\mathrm{d}t
  \;\approx\; \frac{1}{T}\sum_{i=1}^{T} I^{S}_{t_i}(u,v).
  \label{eq:blur}
\end{equation}
Under a rolling shutter with a short exposure, nothing is integrated.
Row $v$ simply begins---and effectively completes---its exposure at
\begin{equation}
  t_{c,n}(v) \;=\; n\Delta \;+\; \delta_c \;+\; \tfrac{v}{H}\,\rho_c,
  \label{eq:rowtime}
\end{equation}
so the observed frame samples a \emph{different} latent sharp image at every row:
\begin{equation}
  I^{\mathrm{RS}}_{c,n}(u,v) \;=\; I^{S}_{t_{c,n}(v)}(u,v).
  \label{eq:rs}
\end{equation}
Both are special cases of one shutter model whose limits are summarized in \cref{tab:shutter}: blur averages over time with a row-independent window, rolling shutter selects a single time that depends on the row.

\begin{table}[t]
  \centering\small
  \setlength{\tabcolsep}{4pt}
  \begin{tabular}{llcc}
    \toprule
    Regime & Condition & Operator & Studied in \\
    \midrule
    Global shutter & $\rho\!=\!0,\ \tau\!\to\!0$ & identity & \cite{hu2024gauhuman,qian20243dgsavatar} \\
    Motion blur    & $\rho\!=\!0,\ \tau\!>\!0$   & averaging & \cite{niu2026madavatar} \\
    Rolling shutter& $\rho\!>\!0,\ \tau\!\to\!0$ & selection & \textbf{ours} \\
    \bottomrule
  \end{tabular}
  \caption{\textbf{Shutter regimes.} Blur and rolling shutter differ in \emph{how} sub-frame renderings are combined, not in what has to be rendered---which is why the avatar framework carries over and only the combination operator changes.}
  \label{tab:shutter}
\end{table}

\paragraph{Discrete 3D-aware model}
Writing the latent sharp image as the rendering of the avatar under the body state at that instant, and discretizing the readout interval into $T$ contiguous row bands
$\mathcal{V}_i=\big\{v: \lceil \tfrac{(i-1)H}{T}\rceil \le v < \lceil \tfrac{iH}{T}\rceil \big\}$
captured at
\begin{equation}
  t_{c,n,i} \;=\; n\Delta + \delta_c + \tfrac{i-1/2}{T}\,\rho_c ,
  \label{eq:bandtime}
\end{equation}
we obtain the model we optimize:
\begin{equation}
  \hat I^{\mathrm{RS}}_{c,n} \;=\; \sum_{i=1}^{T} \mathbf{M}_i \odot
  \rnd\Big(\bW\big(\bG,\ \bS_{t_{c,n,i}}\big);\,\bP_c\Big),
  \label{eq:rs_discrete}
\end{equation}
where $\mathbf{M}_i$ is the binary mask of band $\mathcal{V}_i$ and $\odot$ is the Hadamard product.
Comparing \cref{eq:rs_discrete} with the blur model $\hat I^{B}=\frac{1}{T}\sum_i \rnd(\bW(\bG,\bS_{t_i});\bP_c)$ makes the change explicit: the same $T$ virtual sharp renderings are produced, and the uniform average $\frac{1}{T}\sum_i$ is replaced by the masked sum $\sum_i \mathbf{M}_i\odot$.
Everything else in the pipeline is untouched: the canonical representation, the warping, the rasterizer and the optimizer are those of the blur model.

\paragraph{How many bands}
The residual of \cref{eq:rs_discrete} differs in character from the blur case.
Averaging is a smoothing operation, so a coarse $T$ merely softens the synthesized image and the residual is spatially unstructured.
Band composition instead quantizes the capture time into a piecewise-constant function of $v$, leaving horizontal \emph{seams} at band boundaries displaced by up to $|\dot v|\,\rho_c/(2T)$ pixels---a structured error, which one would expect to force a finer discretization than blur requires.
Empirically it does not.
Sweeping $T\in\{K,2K,4K\}$ at fixed $K$ (\cref{tab:ablation}) changes the result by at most $0.11$\,dB, and does so \emph{monotonically downwards}.
Two effects cap the useful $T$ far below $H$.
The intra-frame trajectory is a degree-$(P\!-\!1)$ polynomial by construction (\cref{eq:spline}), so once $T$ oversamples that polynomial the extra bands merely re-sample the same curve; and since band $i$ is supervised on only $1/T$ of the rows, doubling $T$ halves the gradient each rendering carries while the cost grows linearly.
We therefore set $T\!=\!K$ throughout, and $T$ remains a cost knob rather than an accuracy knob.

\subsection{3D Human Motion Model}
\label{sec:motion}

\cref{eq:rs_discrete} needs $T$ body states inside every frame, which a per-frame SMPL estimate cannot supply.
We adopt the sub-frame motion model of the blur setting~\cite{niu2026madavatar} unchanged.

\paragraph{Sub-frame pose interpolation}
For each frame we keep $P$ learnable control values per joint, $\tilde\Theta^{j}=[\tilde\Theta^{j}_{1},\dots,\tilde\Theta^{j}_{P}]$, and interpolate them with a B-spline of order $P$.
With the normalized time basis
\begin{equation}
  \mathbf{B}(t)=\Big[\,1,\ \tfrac{t}{T},\ \big(\tfrac{t}{T}\big)^{2},\ \dots,\ \big(\tfrac{t}{T}\big)^{P-1}\Big],
  \label{eq:basis}
\end{equation}
the pose of joint $j$ at sub-frame index $t$ is
\begin{equation}
  \hat\Theta^{j}_{t} \;=\; \mathbf{B}(t)\,\mathbf{M}^{P}\,\tilde\Theta^{j},
  \label{eq:spline}
\end{equation}
where $\mathbf{M}^{P}\in\R^{P\times P}$ is the B-spline basis matrix
\begin{equation}
  \mathbf{M}^{P}_{i,j}=\frac{\binom{P-1}{P-1-i}}{(P-1)!}\sum_{s=j}^{P-1}(-1)^{s-j}\binom{P}{s-j}(P\!-\!s\!-\!1)^{P-1-i}.
  \label{eq:Mmatrix}
\end{equation}
The same interpolation is applied to the global orientation and translation.
The $T$ sub-frame indices are placed symmetrically about the frame centre and span a $\gamma$ fraction of the frame interval, per \cref{eq:bandtime}.

\paragraph{Pose deformation}
Skinning from an interpolated pose alone cannot account for clothing and soft-tissue dynamics, so a small network predicts a residual on top of the interpolated pose,
\begin{equation}
  \Delta^{j}_{t}=\mathcal{G}_{\mathrm{disp}}\big(\hat\Theta^{j}_{t};\theta_{\mathrm{disp}}\big),
  \qquad
  \Theta^{j}_{t}=\hat\Theta^{j}_{t}+\Delta^{j}_{t}.
  \label{eq:disp}
\end{equation}

\paragraph{Inter-frame motion regularization}
Control points are defined per frame, so nothing ties the end of one frame's trajectory to the start of the next.
We enforce continuity with a geodesic penalty on $\SOthree$,
\begin{equation}
  \bL_{\mathrm{reg}}=\frac{1}{J(N\!-\!1)}\sum_{n=1}^{N-1}\sum_{j=1}^{J}
  \big\|\hat\Theta^{j}_{n,T}-\hat\Theta^{j}_{n+1,1}\big\|_{G}.
  \label{eq:reg}
\end{equation}
Its role differs from the blur setting.
There, $\bL_{\mathrm{reg}}$ carries a second and heavier burden: the blur operator is invariant to reversing time within the exposure, so the \emph{direction} of the sub-frame motion is unidentifiable without it.
A rolling shutter has no such ambiguity---the scan order is an observable arrow of time---so here $\bL_{\mathrm{reg}}$ only has to enforce continuity, and it is that lighter role we ablate in \cref{sec:ablation}.

\paragraph{Shape and skinning refinement}
The shape $\hat\beta$ is a single vector optimized from its initial estimate and shared by the whole sequence, and the skinning weights are refined as $\hat{\mathcal{B}}=\tilde{\mathcal{B}}+\delta$ with $\delta$ predicted from the canonical Gaussian coordinates.

\subsection{Joint Optimization}
\label{sec:optim}

All quantities---the canonical Gaussians, the per-frame control points, the deformation network, the shape and the skinning weights---are optimized jointly against the raw RS observations by minimizing
\begin{equation}
  \bL \;=\; \big\|\hat I^{\mathrm{RS}}-I^{\mathrm{RS}}\big\|_{1} \;+\; \lambda\,\bL_{\mathrm{reg}} ,
  \label{eq:loss}
\end{equation}
with $\hat I^{\mathrm{RS}}$ given by \cref{eq:rs_discrete}.
Control points are initialized from an off-the-shelf per-frame SMPL fit~\cite{easymocap} with small random perturbations, and adaptive density control follows GauHuman~\cite{hu2024gauhuman}.
Note that a per-frame fit to RS frames is biased towards the pose at the \emph{middle} band's capture time, which makes it a reasonable initialization for the spline but not a usable final answer.

\section{Experiments}
\label{sec:exp}

\subsection{Benchmark}
\label{sec:datasets}

No dataset exists for rolling-shutter human avatar reconstruction, so we build \dsyn.
We start from the refined ZJU-MoCap~\cite{peng2021neuralbody} sequences and use the six subjects (\texttt{377}, \texttt{386}, \texttt{387}, \texttt{392}, \texttt{393}, \texttt{394}) at their native $1024\!\times\!1024$ resolution, with $556$--$859$ frames each.
Following the protocol used to build the blur benchmark~\cite{niu2026madavatar}, a group of $K$ consecutive source frames becomes one output frame; $K$ is thus a frame-rate divider, and we use $K\!=\!11$.
We set $\gamma\!=\!1$ throughout, so the readout occupies the entire frame period and the last row of a frame is captured just as the first row of the next one begins---the continuously reading sensor of \cref{tab:shutter}.
The readout window of output frame $n$ is therefore $[\,\mathrm{mid}-K/2,\ \mathrm{mid}+K/2\,]$ in source-frame units, with $\mathrm{mid}=nK+\lfloor K/2\rfloor$, and $K$ alone sets how much body motion a single frame contains.

Synthesizing this needs the scene at sub-frame resolution, so every adjacent source pair the window touches is temporally upsampled $2^{e}$ times with RIFE~\cite{huang2022rife}; the RS frame is then assembled by taking each row from the sub-frame instant whose timestamp matches \cref{eq:rowtime}.
We choose $e$ from the seam bound of \cref{sec:formation}: measuring $D$, the image-space displacement of the fastest body point over a whole readout window, by optical flow over all subjects and views gives $D\!=\!44$--$60$\,px, and we take the smallest $e$ keeping the band-boundary discontinuity $D/(K2^{e})$ below half a pixel.
This yields $e\!=\!4$, \ie{} $177$ sub-frame instants per frame and a seam of $0.34$\,px.
That is far finer than a blur benchmark would need---averaging hides a coarse discretization and band composition does not---and it is the concrete price of the structured error discussed in \cref{sec:formation}.

Two details make the benchmark cleaner.
We centre the readout window on the middle source frame of a group rather than starting it at the group boundary, so the mid-row capture time coincides with a \emph{real} captured frame and the evaluation ground truth is an original sharp image rather than an interpolation.
Because a centred window with $\gamma\!=\!1$ reaches half an interval beyond the group on each side, the first group of every sequence is dropped, leaving $354$ frames per view summed over the six subjects.
Four cameras (\texttt{04}, \texttt{10}, \texttt{16}, \texttt{22}) are used for training and, following the ZJU-MoCap protocol, camera \texttt{03} is excluded, leaving $18$ held-out cameras for novel-view evaluation; ground truth is the original global-shutter frame at the reference time $t_{c,n}(H/2)$.

\paragraph{Metrics}
We report PSNR, SSIM~\cite{wang2004ssim} and LPIPS~\cite{zhang2018lpips} on held-out views, computed inside the dilated body mask.

\subsection{Baselines}
\label{sec:baselines}
\begin{enumerate}[label=(\alph*),leftmargin=1.6em,itemsep=1pt,topsep=2pt]
  \item \textbf{Shutter-oblivious.} GauHuman~\cite{hu2024gauhuman} trained directly on RS frames. This is what a practitioner does today.
  \item \textbf{Two-stage.} A 2D RS correction network applied per view, followed by GauHuman. We take CVR~\cite{fan2022cvr} and the more recent JAMNet~\cite{fan2023jamnet}. Both synthesize a global-shutter frame at a selectable reference time; we set it to the middle scanline, matching our evaluation target, and select each method's checkpoint by measured accuracy on our data rather than by its default.
  \item \textbf{Blur-model transfer.} MAD-Avatar~\cite{niu2026madavatar} applied to RS input, testing whether a motion-aware \emph{blur} model suffices.
\end{enumerate}

\subsection{Implementation Details}
\label{sec:impl}
\ours{} is built on the GauHuman~\cite{hu2024gauhuman} codebase.
The rasterizer is used unmodified; the only change to the training loop is that the $T$ sub-frame renderings are combined by \cref{eq:rs_discrete} instead of being averaged.
We use $P\!=\!4$ control points, $T\!=\!K$ row bands (\cref{tab:ablation}), $\lambda\!=\!1$, and train for $20$k iterations with Adam.
Training takes roughly $55$ minutes per subject on a single RTX 4090.
At equal $T$ this is indistinguishable from the blur model: both render the same $T$ sub-frames, which dominates the cost, and replacing the average by the masked sum adds no rendering.
Full hyperparameters and per-subject numbers are in the supplement.

\subsection{Results}
\label{sec:res_syn}

\begin{table}[t]
  \centering\footnotesize
  \setlength{\tabcolsep}{3.5pt}
  \begin{tabular}{ll ccc}
    \toprule
    Family & Method & PSNR$\uparrow$ & SSIM$\uparrow$ & LPIPS$\downarrow$ \\
    \midrule
    Shutter-oblivious
      & GauHuman~\cite{hu2024gauhuman}        & 23.69 & 0.803 & 0.153 \\
    \midrule
    \multirow{2}{*}{Two-stage}
      & CVR~\cite{fan2022cvr} + GH            & 23.99 & 0.809 & 0.153 \\
      & JAMNet~\cite{fan2023jamnet} + GH      & 23.96 & 0.808 & 0.155 \\
    \midrule
    Motion-aware
      & MAD-Avatar~\cite{niu2026madavatar}    & 22.99 & 0.789 & 0.170 \\
    \midrule
      & \textbf{\ours{} (ours)}               & \textbf{24.68} & \textbf{0.819} & \textbf{0.138} \\
    \bottomrule
  \end{tabular}
  \caption{\textbf{Novel-view synthesis on \dsyn} at $K\!=\!11$, averaged over the $18$ held-out cameras of each subject and then over the six subjects (frame-count weighting moves every entry by less than $0.03$).
  Note that the blur model scores \emph{below} the shutter-oblivious baseline.}
  \label{tab:syn_main}
\end{table}

\begin{figure*}[t]
  \centering
  \includegraphics[width=\textwidth]{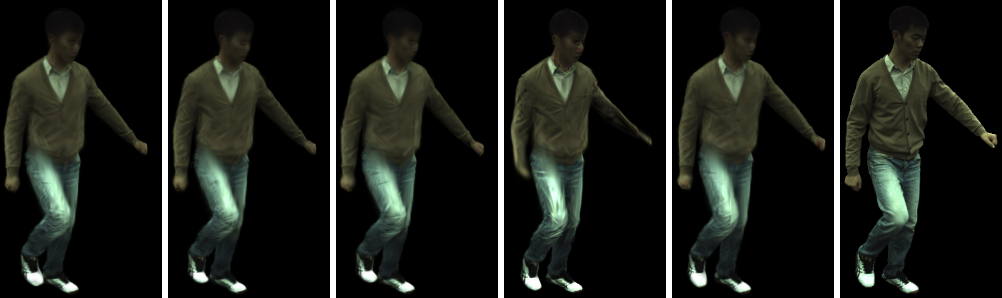}\\[2pt]
  \makebox[0.1667\textwidth]{\footnotesize GauHuman}%
  \makebox[0.1667\textwidth]{\footnotesize CVR + GH}%
  \makebox[0.1667\textwidth]{\footnotesize JAMNet + GH}%
  \makebox[0.1667\textwidth]{\footnotesize MAD-Avatar}%
  \makebox[0.1667\textwidth]{\footnotesize \textbf{\ours{} (ours)}}%
  \makebox[0.1667\textwidth]{\footnotesize GT}
  \caption{\textbf{Novel-view synthesis on \dsyn} at $K\!=\!11$, held-out camera.
  The outstretched forearm sweeps the largest image-space distance within one readout, and is where the methods differ.}
  \label{fig:qual_syn}
\end{figure*}

\cref{tab:syn_main} reports novel-view synthesis, and \cref{fig:qual_syn} the corresponding qualitative behavior.
The comparison separates three questions: whether the shutter must be modeled at all (row 1 \vs{} ours), whether modeling it in 2D is sufficient (rows 2--3), and whether an existing motion-aware model of a \emph{different} corruption transfers (row 4).

Modeling the shutter is worth $0.99$\,dB over training on the RS frames as if they were instantaneous, and \ours{} is ahead on all six subjects.
\cref{fig:qual_syn} shows where the difference sits.
Every method recovers the torso and they separate on the arm: the shutter-oblivious reconstruction blurs it into the background and rounds the hand into a blob, the two 2D-corrected variants sharpen the torso but leave the forearm thickened with a soft contour, and the blur model leaves a dark streak along its length, which is what averaging renderings the sensor never averaged produces.
Ours keeps the forearm's taper and the gap between hand and hip.
None of them recovers the cardigan buttons; those are absent for a different reason, being a limit of reconstructing a canonical avatar from four views rather than a shutter effect, and they bound every row of \cref{tab:syn_main} equally.

The blur model is worth a closer look, because MAD-Avatar shares the sub-frame apparatus this method depends on: the same pose spline, the same $T$ renderings per frame, the same optimizer.
It differs only in combining those renderings by a uniform average rather than by scanline composition, and that costs $1.69$\,dB against ours and $0.70$\,dB against the shutter-oblivious baseline, which models no sub-frame structure at all.
Falling behind that baseline is the part worth explaining.
Averaging $T$ renderings taken at different instants produces a synthetic image in which every row is a mixture, and it is then matched against an observation in which every row is a single sample; the residual is nonzero everywhere and the optimizer has no way to attribute it, so it distributes the error over the canonical Gaussians.
The shutter-oblivious model makes a cruder assumption but a self-consistent one, and is penalised only where the body actually moved.

The two correctors both improve the images they produce.
Scored against the sharp reference inside the body region, before any avatar is trained, CVR gains $0.84$\,dB and JAMNet $0.97$\,dB.
Only about a third of that reaches the avatar, $0.30$ and $0.27$\,dB, leaving both roughly $0.7$\,dB behind modeling the shutter directly.
The shortfall belongs to the pipeline rather than to either network: each view is corrected in isolation, nothing constrains the corrected views to agree with one another in 3D, and the reconstruction downstream cannot separate a residual correction error from appearance.
CVR carries a further cost that JAMNet does not, which makes the mechanism visible.
Trained on driving scenes where the distortion comes from camera motion, it estimates a single flow field and resamples the whole image; on a static rig observing a moving subject that is the wrong prior for most pixels, and its full-frame PSNR drops by $4.0$\,dB even as the body region improves, whereas JAMNet leaves the frame essentially intact ($+0.2$\,dB).
The two nonetheless deliver almost the same avatar, which suggests that the quality of the per-view restoration is not what bounds this pipeline.
Modeling the readout inside the image formation process avoids the trade, since the shutter is then explained once, in the geometry the views already share.

\subsection{Ablation}
\label{sec:ablation}

\cref{tab:ablation} removes one component at a time.
Replacing the intra-frame trajectory with a single pose costs $3.25$\,dB, twice what removing the rolling-shutter model alone costs.
The two are not independent, since a constant intra-frame pose leaves the operator nothing to select between; this row therefore bounds their combined contribution rather than isolating the trajectory's.
Quadrupling the number of bands moves the result by $0.11$\,dB, and downwards.
The intra-frame trajectory is a cubic (\cref{eq:spline}) that $T\!=\!K$ already oversamples, so the extra bands re-sample the same curve, and each rendering is supervised on only $1/T$ of the rows, so a larger $T$ also thins the gradient it carries.
Of the remaining components, the inter-frame prior is worth $1.39$\,dB and the pose deformation network $0.65$\,dB, while the LBS refinement contributes $0.08$\,dB and could be dropped.

\begin{table}[t]
  \centering\footnotesize
  \setlength{\tabcolsep}{4pt}
  \begin{tabular}{lccc}
    \toprule
    Variant & PSNR$\uparrow$ & SSIM$\uparrow$ & LPIPS$\downarrow$ \\
    \midrule
    \textbf{Full model} ($T\!=\!K$)             & \textbf{24.65} & \textbf{0.858} & \textbf{0.096} \\
    \midrule
    w/o RS model ($\rho\!\equiv\!0$)           & 23.08 & 0.835 & 0.114 \\
    w/o B-spline (one pose per frame)          & 21.41 & 0.794 & 0.176 \\
    w/o pose deformation                       & 24.00 & 0.847 & 0.109 \\
    w/o LBS refinement                         & 24.57 & 0.855 & 0.098 \\
    w/o $\bL_{\mathrm{reg}}$                   & 23.26 & 0.837 & 0.109 \\
    \midrule
    \multicolumn{4}{l}{\textit{Row bands (cost $2\!\times$ and $4\!\times$ the full model)}} \\
    $T\!=\!2K$                                 & 24.59 & 0.857 & 0.096 \\
    $T\!=\!4K$                                 & 24.54 & 0.856 & 0.096 \\
    \bottomrule
  \end{tabular}
  \caption{\textbf{Ablations} on \dsyn{} at $K\!=\!11$, subject \texttt{377}, which is why the full-model row differs from the six-subject average in \cref{tab:syn_main}.
  Evaluation renders one body state at the mid-row capture time and is independent of $T$, so the last block is directly comparable to the full model.}
  \label{tab:ablation}
\end{table}

\subsection{Limitations}
\label{sec:limitations}
Four limitations are worth stating.
First, the intra-frame trajectory is a degree-$(P\!-\!1)$ polynomial, so motion a cubic cannot follow within a single frame lies outside the model, and---as \cref{tab:ablation} shows---no amount of extra readout discretization recovers it.
Second, we operate in sRGB and assume a well-exposed capture, so the method has no account of low-light degradation, where the sensor noise and the tone curve interact with the readout~\cite{niu2023visibility,niu2023nir,niu2023physics,guo2020zero,mildenhall2022nerf}.
Third, sub-frame motion is carried entirely by SMPL, so anything without a corresponding joint---a handheld object, a loose garment---has no way to move within a frame, and its scanline-dependent displacement cannot be recovered; adaptive or learned joints~\cite{zhan2025towards} together with non-rigid modelling are a natural route to lifting this.
Finally, our evaluation is synthetic: the readout schedule is exact by construction, whereas a real sensor's is only approximately known.
Beyond these, generative priors~\cite{niu2024mofa,zhao2024cv,zhao2024stereocrafter,liu20243dgs,wu2025difix3d,niu2025anicrafter,chen2024mvsplat360} could plausibly refine regions that four views leave underconstrained---the cardigan buttons of \cref{fig:qual_syn} being one example---though doing so would trade measurement for hallucination and is out of scope here.

\section{Conclusion}
\label{sec:conclusion}

We have argued that the rolling shutter belongs in the image formation model of animatable human avatars rather than in a preprocessing stage.
Placing it there changes what a frame is taken to represent: a schedule rather than a snapshot, with the body posed once per scanline instead of once per image.
Little else has to change, because a motion-aware avatar already renders the body at several sub-frame instants; going from blur to rolling shutter only replaces the uniform average of those renderings with a scanline-wise composition.

One conclusion generalizes beyond our system: corruption models are not interchangeable even when their machinery is.
Blur and rolling shutter differ only in how sub-frame renderings are combined, yet blur discards the direction of sub-frame motion while a rolling shutter records it, so a prior written to repair one is doing a different job in the other.
We measure this as a blur model that, given the identical spline, renderings and optimizer, scores below a baseline that models no sub-frame structure at all.

One question we do not answer is worth stating.
Because scanlines across a rig tile the time axis densely, an RS capture samples the body's motion at more distinct instants than a global-shutter capture at the same frame rate.
Whether that latent sub-frame information can be turned into usable temporal super-resolution is something our formulation makes it possible to ask, and we leave it open.

Natural extensions include event- or IMU-assisted initialization for very fast motion, multi-person capture where scanline schedules interact with inter-person occlusion, and monocular RS avatar reconstruction, where the reduced multi-view redundancy will call for additional priors.

{
    \small
    \bibliographystyle{ieeenat_fullname}
    \bibliography{main}
}


\end{document}